\documentclass[10pt,twocolumn,letterpaper]{article}

\usepackage{cvpr}
\usepackage{times}
\usepackage{epsfig}
\usepackage{graphicx}
\usepackage{amsmath}
\usepackage{amssymb}

\usepackage{subfigure}
\usepackage{booktabs} 
\usepackage{multirow}
\usepackage{url}
\usepackage{enumitem}
\usepackage{microtype}
\usepackage[font=small,labelfont=bf]{caption}
\usepackage{xcolor}
\usepackage{diagbox}
\usepackage{nameref}

\usepackage[pagebackref=true,breaklinks=true,letterpaper=true,colorlinks,bookmarks=false]{hyperref}

\cvprfinalcopy % *** Uncomment this line for the final submission

\def\cvprPaperID{4898} % *** Enter the CVPR Paper ID here

\newcommand{\norm}[1]{\left\lVert#1\right\rVert}
\newcommand{\Paragraph}[1]{\vspace{0.5mm} \noindent \textbf{#1.} \hspace{0mm}}
\newcommand{\Section}[1]{\vspace{-2mm} \section{#1} \vspace{-1.5mm}}
\newcommand{\SubSection}[1]{\vspace{-1mm} \subsection{#1} \vspace{-1.5mm}}
\newcommand{\SubSubSection}[1]{\vspace{-3mm} \subsubsection{#1} \vspace{-2mm}}

\def\L{\mathcal{L}}
\def\Enc{\mathcal{E}}
\def\Dec{\mathcal{D}}
\def\Lxrecon{\L_{\text{xrecon}}}
\def\Lid{\L_{\text{id-inc-avg}}}
\def\fa{\mathbf{f}_a}
\def\fg{\mathbf{f}_g}

\ifcvprfinal\fi

\begin{document}

%%%%%%%%% TITLE
\title{Gait Recognition via Disentangled Representation Learning}

\author{Ziyuan Zhang, Luan Tran, Xi Yin, Yousef Atoum, Xiaoming Liu\\
Michigan State University\\
% 428 South Shaw Lane, East Lansing, MI 48824\\
{\tt\small \{zhang835, tranluan, yinxi1, atoumyou, liuxm\}@msu.edu}
% For a paper whose authors are all at the same institution,
% omit the following lines up until the closing ``}''.
% Additional authors and addresses can be added with ``\and'',
% just like the second author.
% To save space, use either the email address or home page, not both
\and
Jian Wan, Nanxin Wang\\
Ford Research and Innovation Center\\
% 2101 Village Rd, Dearborn, MI 48124\\
{\tt\small \{jwan1, nwang1\}@ford.com}
}

\maketitle

\thispagestyle{empty}

%------------------------------------------------------------------------
%%%%%%%%% ABSTRACT
\begin{abstract}
\vspace{-2mm}
  Gait, the walking pattern of individuals, is one of the most important biometrics modalities. 
  Most of the existing gait recognition methods take silhouettes or articulated body models as the gait features. 
  These methods suffer from degraded recognition performance when handling confounding variables, such as clothing, carrying and view angle. 
  To remedy this issue, we propose a novel AutoEncoder framework to explicitly disentangle pose and appearance features from RGB imagery and the LSTM-based integration of pose features over time produces the gait feature. 
  In addition, we collect a Frontal-View Gait (FVG) dataset to focus on gait recognition from frontal-view walking, which is a challenging problem since it contains minimal gait cues compared to other views. 
  FVG also includes other important variations, e.g., walking speed, carrying, and clothing. 
  With extensive experiments on CASIA-B, USF and FVG datasets, our method demonstrates superior performance to the state of the arts quantitatively, the ability of feature disentanglement qualitatively, and promising computational efficiency.
\end{abstract}

\begin{figure}[h!]
\includegraphics[width=8cm]{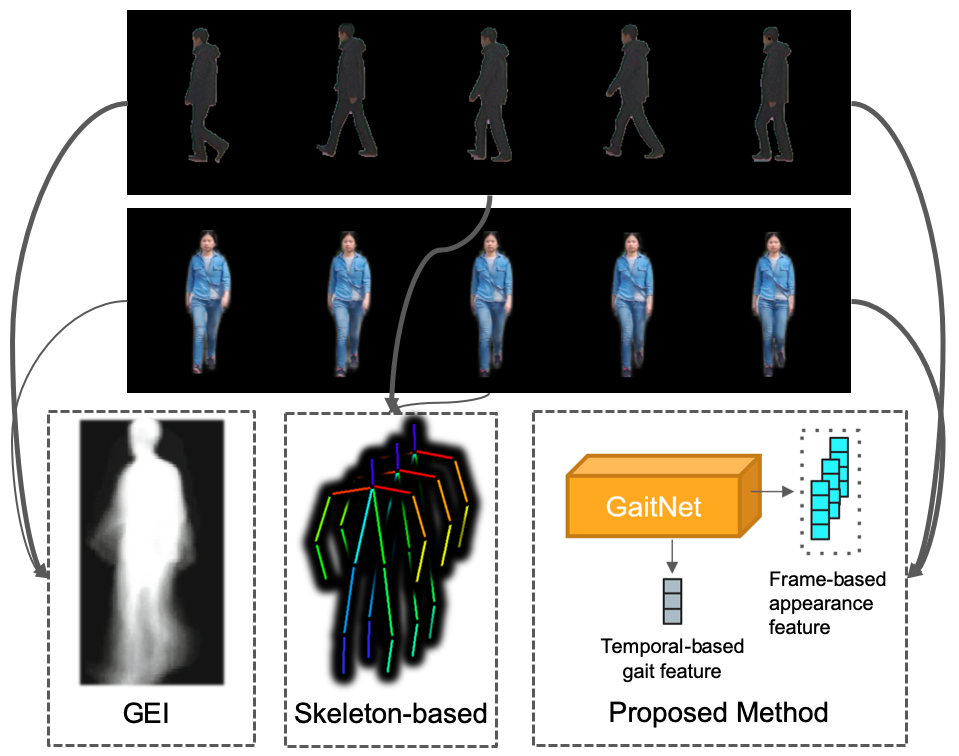}
\centering
\caption{We propose a novel CNN-based model, termed GaitNet, to automatically learn the disentangled gait feature from a walking video, as opposed to handcrafted GEI, or skeleton-based features.
While many conventional gait databases study side-view imagery, we collect a new gait database where both gallery and probe are captured in frontal-views.}
\label{fig:concept}
\vspace{-4mm}
\end{figure}

%------------------------------------------------------------------------
%%%%%%%%% INTRODUCTION
\Section{Introduction}
Biometrics measures people's unique physical and behavioral characteristics to recognize the identity of an individual. 
Gait~\cite{nixon2010human}, the walking pattern of an individual, is one of the biometrics modalities, e.g., face, fingerprint, and iris.
Gait recognition has the advantage that it can operate at a distance without user cooperation.
Also, it is difficult to camouflage. 
Due to these advantages, gait recognition is applicable to many applications such as person identification, criminal investigation, and healthcare.

As other recognition problems in vision, the core of gait recognition lies in extracting {\it gait-related features} from the video frames of a walking person, where the prior approaches are categorized into two types: appearance-based and model-based methods.
The appearance-based methods such as Gait Energy Image (GEI)~\cite{han2006individual} take the averaged silhouette image as the gait feature. 
While having a low computational cost and can handle low-resolution imagery, it can be sensitive to variations such as clothes change, carrying, view angles and walking speed~\cite{sarkar2005humanid,bashir2009gait,wu2017comprehensive,bashir2010gait,
hossain2010clothing,alotaibi2017improved}.
The model-based method first performs pose estimation and takes articulated body skeleton as the gait feature. 
It shows more robustness to those variations but at a price of a higher computational cost and dependency on pose estimation accuracy~\cite{feng2016learning,ariyanto2012marionette}.
%--------------------------------3-----------------------------------------
\begin{table*}[t!]
\centering
\caption{Comparison of existing gait databases and our collected FVG database.}
\vspace{-2mm}
\label{tab:db}
\resizebox{1\linewidth}{!}{

%\begin{tabular}{|l|l|l|c|c|c|l|}
%\hline
%Dataset & \#Subjects & \#Videos & Environment & Resolution & Format & Variations \\ \hline
%CASIA-B & $124$ & $13,640$ & Indoor & $320$x$240$  & RGB & \begin{tabular}[c]{@{}c@{}}View, Clothing, Carrying \end{tabular} \\ \hline
%USF & $122$ & $1,870$ & Outdoor & $720$x$480$  & RGB & \begin{tabular}[c]{@{}c@{}}View, Ground Surface, Shoes, Carrying, Time \end{tabular} \\ \hline
%OU-ISIR-LP & $4,007$ & $-$ & Indoor & $640$x$480$  & Silhouette & \begin{tabular}[c]{@{}c@{}} View \end{tabular} \\ \hline
%OU-ISIR-LP-Bag & $62,528$ & $-$ & Indoor & $1,280$x$980$  & Silhouette & \begin{tabular}[c]{@{}c@{}} Carrying  \end{tabular} \\ \hline
%FVG (our) & $226$ & $2,856$ & Outdoor & $1,920$x$1,080$  & RGB & \begin{tabular}[c]{@{}c@{}}View, Walking Speed, Carrying, Clothing, Background, Time \end{tabular} \\ \hline
%\end{tabular}
%}
%\vspace{-4mm}
%\end{table*}

\begin{tabular}{lcccccl}
\toprule
Dataset & \#Subjects & \#Videos & Environment & Resolution & Format & Variations \\ \midrule
CASIA-B & $124$ & $13,640$ & Indoor & $320{\times}240$  & RGB & \begin{tabular}[c]{@{}c@{}}View, Clothing, Carrying \end{tabular} \\ %\hline
USF & $122$ & $1,870$ & Outdoor & $720{\times}480$  & RGB & \begin{tabular}[c]{@{}c@{}}View, Ground Surface, Shoes, Carrying, Time \end{tabular} \\ %\hline
OU-ISIR-LP & $4,007$ & $-$ & Indoor & $640{\times}480$  & Silhouette & \begin{tabular}[c]{@{}c@{}} View \end{tabular} \\ %\hline
OU-ISIR-LP-Bag & $62,528$ & $-$ & Indoor & $1,280{\times}980$  & Silhouette & \begin{tabular}[c]{@{}c@{}} Carrying  \end{tabular} \\ %\hline
FVG (ours) & $226$ & $2,856$ & Outdoor & $1,920{\times}1,080$  & RGB & \begin{tabular}[c]{@{}c@{}}View, Walking Speed, Carrying, Clothing, Background, Time \end{tabular} \\ \bottomrule
\end{tabular}
}
\vspace{-4mm}
\end{table*}

%------------------------------------------------------------------------

It is understandable that the challenge in designing a gait feature is the necessity of being invariant to the appearance variation due to clothing, viewing angle, carrying, etc.
Therefore, our desire is to {\it disentangle} the gait feature from the visual appearance of the walking person.
For both appearance-based or model-based methods, such disentanglement is achieved by manually handcrafting the GEI or body skeleton, since neither has color information.
However, we argue that these manual disentanglements may lose certain or create redundant gait information.
E.g., GEI learns the average contours over time, but not the dynamic of how body parts move.
For body skeleton, under carrying condition, certain body joints such as hands may have fixed positions, and hence are redundant information to gait. 

To remedy the issues in handcrafted features, as shown in Fig.~\ref{fig:concept}, this paper aims to automatically disentangle the pose/gait features from appearance features, and use the former for gait recognition.
This disentanglement is realized by designing an autoencoder-based CNN, GaitNet, with novel loss functions.
For each video frame, the encoder estimates two latent representations, pose feature (i.e., frame-based gait feature) and appearance feature, by employing two loss functions: 1) cross reconstruction loss enforces that the appearance feature of one frame, fused with the pose feature of another frame, can be decoded to the latter frame; 2) gait similarity loss forces a sequence of pose features extracted from a video sequence, of the same subject to be similar even under different conditions.
Finally, the pose features of a sequence are fed into a multi-layer LSTM with our designed incremental identity loss to generate the sequence-based gait feature, where two of which can use the cosine distance as the video-to-video similarity metric.

Furthermore, most prior work~\cite{han2006individual,wu2017comprehensive,makihara2017joint,chen2018multi,ariyanto2012marionette,bobick2001gait,cunado2003automatic} often choose the walking video of the side view, which has the richest gait information, as the gallery sequence. However, practically other view angles, such as the frontal view, can be very common when pedestrians toward or away from the surveillance camera. Also, the prior work~\cite{sivapalan2011gait,chattopadhyay2014pose,chattopadhyay2014frontal,nambiar2012frontal} that focuses on frontal view are often based on RGB-D videos, which have richer depth information than RGB videos. 
Therefore, to encourage gait recognition from the frontal-view RGB videos that generally has the minimal amount of gait information, we collect a high-definition (HD,$1080$p) frontal-view gait database with a wide range of variations. It has three frontal-view angles where the subject walks from left $45^{\circ}$, $0^{\circ}$, and right $45^{\circ}$  off the optical axes of the camera. For each of three angles, different variants are explicitly captured including walking speed, clothing, carrying, clutter background, etc.  

The contributions of this work are the following:

1) We propose an autoencoder-based network, GaitNet, with novel loss functions to explicitly disentangle the pose features from visual appearance and use multi-layer LSTM to obtain aggregated gait feature.

2) We introduce a frontal-view gait database, named FVG, including various variations of viewing angles, walking speeds, carrying, clothing changes, background and time gaps. This is the first HD gait database, with a nearly doubled number of subjects than prior RGB gait databases.

3) Our proposed method outperforms state of the arts on three benchmarks, CASIA-B, USF, and FVG datasets.

%-------------------------------------------------------------------------

\Section{Related Work}
\label{sec:relatedwork}

\begin{figure*}[t!]

\includegraphics[trim=0 0 0 0, clip, width=0.9\linewidth]{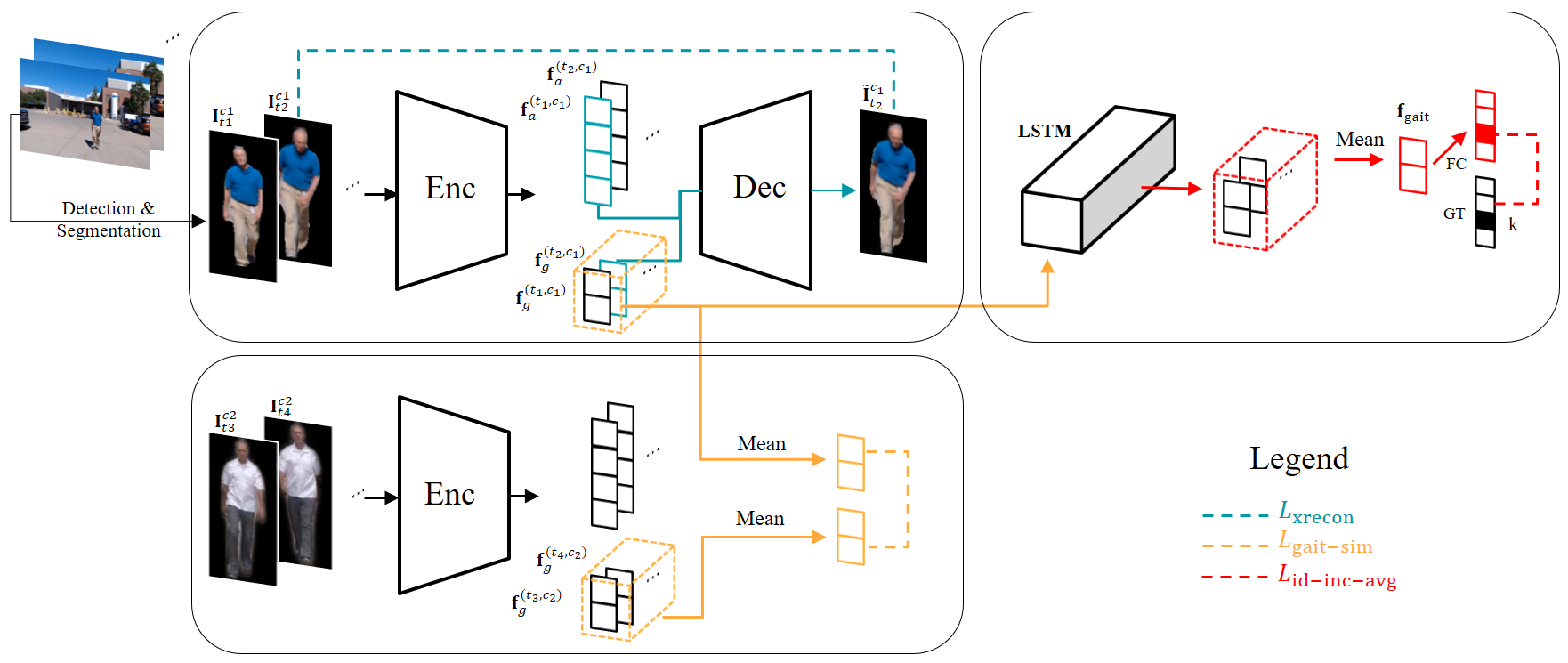}
\centering
\caption{Overall architecture of our proposed approach, with three novel loss functions.}
\label{fig-arc}
\vspace{-4mm}
\end{figure*}
%-------------------------------------------------------------------------

\Paragraph{Gait Representation}
Most prior works are based on two types of gait representations. 
In appearance-based methods, gait energy image (GEI)~\cite{han2006individual} or gait entropy image (GEnI)~\cite{bashir2009gait} are defined by extracting silhouette masks. 
Specifically, GEI uses an averaged silhouette image as the gait representation for a video.
These methods are popular in the gait recognition community for their simplicity and effectiveness.  However, they often suffer from sizeable intra-subject appearance changes due to covariates such as clothing, carrying, views, and walking speed. 
On the other hand, model-based methods~\cite{feng2016learning} fit articulated body models to images and extract kinematic features such as $2$D body joints. 
While they are robust to some covariates such as clothing and speed, they require a relatively higher image resolution for reliable pose estimation and higher computational costs.

In contrast, our approach learns gait information from raw RGB video frames which contain the richer information, thus with higher potential of extracting discriminative gait features.
The most relevant work to ours is~\cite{chen2018multi}, which learns gait features from RGB images via Conditional Random Field. 
Compared to ~\cite{chen2018multi}, our CNN-based approach has the advantage of being able to leverage a large amount of training data and learning more discriminative representation from data with multiple covariates.
This is demonstrated by our extensive comparison with~\cite{chen2018multi} in Sec.~\ref{sec:CASIA-B}.

\Paragraph{Gait Databases}
There are many classic gait databases such as SOTON Large dataset~\cite{shutler2004large}, USF~\cite{sarkar2005humanid}, CASIA-B~\cite{hofmann2014tum}, OU-ISIR~\cite{makihara2012isir}, TUM GAID~\cite{hofmann2014tum} and etc. 
We compare our FVG database with the most widely used ones in Tab.~\ref{tab:db}.
CASIA-B is a large multi-view gait database with three variations: view angle, clothing, and carrying.
Each subject is captured from $11$ views under three conditions: normal walking (NM), walking in coats (CL) and walking while carrying bags (BG). 
For each view, $6$, $2$ and $2$ videos are recorded from normal, coats and bags conditions.
USF database has $122$ subjects with five variations, totaling $32$ conditions for each subject. 
It contains two view angles (left and right), two ground surface (grass and concrete), shoes change, carrying condition and time. 
While OU-ISIR-LP and OU-ISIR-LP-Bag are large datasets, we can not leverage them as only the  silhouette is publicly released.

Unlike those databases, our FVG database focuses on the frontal view, with $3$ different near frontal-view angles towards the camera, and other variations including walking speed, carrying, clothing, cluttered background and time.

\Paragraph{Disentanglement Learning}
Besides model-based approaches~\cite{tran2018nonlinear, tran2019towards,disentangling-features-in-3d-face-shapes-for-joint-face-reconstruction-and-recognition} representing data with semantic latent vectors; data-driven disentangled representation learning approaches are gaining popularity in computer vision community.
DrNet~\cite{denton2017unsupervised} disentangles content and pose vectors with a two-encoders architecture, which removes content information in the pose vector by generative adversarial training. 
The work of~\cite{balakrishnan2018synthesizing} segments foreground masks of body parts by $2$D pose joints via U-Net~\cite{ronneberger2015u} and then transforms body parts to desired motion with adversarial training.
Similarly, \cite{esser2018variational} utilizes U-net and Variational Auto Encoder (VAE) to disentangle an image into appearance and shape.
DR-GAN~\cite{tran2017disentangled,tran2018representation} achieves state-of-the-art performances on pose-invariant face recognition by explicitly disentangling pose variation with a multi-task GAN~\cite{goodfellow2014generative}. 

Different from~\cite{denton2017unsupervised,balakrishnan2018synthesizing,esser2018variational}, our method has only one encoder to disentangle the appearance and gait information, through the design of novel loss functions without the need for adversarial training. 
Unlike DR-GAN~\cite{tran2018representation}, our method does not require adversarial training, which makes training more accessible.
Further, pose labels are used in DR-GAN training so as to disentangle identity feature from the pose. 
However, to disentangle gait and appearance feature from the RGB information, there is no gait nor appearance {\it label} to be utilized for our method, since the type of walking pattern or clothes cannot be defined as discrete classes.

\Section{Proposed Approach}
Let us start with a simple example. 
Assuming there are three videos, where videos $1$ and $2$ capture subject A wearing t-shirt and long down coat respectively, and in video $3$ subject B wears the same long down coat as in video $2$.
The objective is to design an algorithm, from which the gait features of video $1$ and $2$ are the same, while those of video $2$ and $3$ are different.
Clearly, this is a challenging objective, as the long down coat can easily dominate the feature extraction, which would make videos $2$ and $3$ to be more similar than videos $1$ and $2$ in the latent space of gait features.
Indeed the core challenge, as well as the objective, of gait recognition is to extract gait features that are discriminative among subjects, but invariant to different confounding factors, such as viewing angles, walking speeds and appearance. %carrying conditions, clothing, and

Our approach to achieve this objective is via feature disentanglement - separating the gait feature from appearance information for a given walking video. 
As shown in Fig.~\ref{fig-arc},
the input to our model is a video frame, with background removed using any off-the-shelf pedestrian detection and segmentation method~\cite{he2017mask,illuminating-pedestrians-via-simultaneous-detection-segmentation,pedestrian-detection-with-autoregressive-network-phases}. An encoder-decoder network, with carefully designed loss functions, is used to disentangle the appearance and pose features for each video frame. Then, a multi-layer LSTM explores the temporal dynamics of pose features and aggregates them to a sequence-based gait feature for the identification purpose.
In this section, we first present the feature disentanglement, followed by temporal aggregation, and finally implementation details.

%-------------------------------------------------------------------------
\SubSection{Appearance and Pose Feature Disentanglement}
\label{sec:disentangle}

For the majority of gait recognition datasets, there is a limited appearance variation within each subject. Hence, appearance could be a discriminate cue for identification during training as many subjects can be easily distinguished by their clothes. Unfortunately, any networks or feature extractors relying on appearance will not generalize well on the test set or in practice, due to potentially diverse clothing or appearance between two videos of the same subject.

This limitation on training sets also prevents us from learning good feature extractors if solely relying on identification objective.
Hence we propose to learn to disentangle the gait feature from the visual appearance in an unsupervised manner.
Since a video is composed of frames, disentanglement should be conducted on the frame level first.
Because there is no dynamic information within a video frame, we aim to disentangle the pose feature from the visual appearance for a frame.
The dynamics of pose features over a sequence will contribute to the gait feature.
In other words, we view the pose feature as the manifestation of video-based gait feature at a specific frame.

To this end, we propose to use an encoder-decoder network architecture with carefully designed loss functions to disentangle the pose feature from appearance feature. 
The encoder, $\Enc$, encodes a feature representation of each frame, $\mathbf{I}$, and explicitly splits it into two parts, namely appearance $\mathbf{f}_{a}$ and pose $\mathbf{f}_{g}$ features:
\begin{equation} \label{eq1}
    \fa,\fg = \Enc(\mathbf{I}).
\end{equation}
These two features are expected to fully describe the original input image. As they can be decoded back to the original input through a decoder $\Dec$:
\begin{equation} \label{eq3}
\widetilde{\mathbf{I}} = \Dec(\fa,\fg).
\end{equation}
We now define the various loss functions defined for learning the encoder, $\Enc$, and decoder $\Dec$.

\Paragraph{Cross Reconstruction Loss}
The reconstructed $\widetilde{\mathbf{I}}$ should be close to the original input $\mathbf{I}$.
However, enforcing self-reconstruction loss as in typical auto-encoder can't ensure the appearance $\fa$ learning appearance information across the video and $\fg$ representing pose information in each frame. Hence we propose the cross reconstruction loss, using an appearance feature $\fa^{t_1}$ of one frame and pose feature $\fg^{t_2}$  of another one to reconstruct the latter frame:
\begin{equation} \label{eq4}
\Lxrecon = \norm{ \Dec(\fa^{t_1},\fg^{t_2})-\mathbf{I}_{t_2} }^2_2,
\end{equation}
where $\mathbf{I}_{t}$ is the video frame at the time step $t$.

The cross reconstruction loss, on one hand, can play a role as the self-reconstruction loss to make sure the two features are sufficiently representative to reconstruct video frames. On the other hand, as we can pair a pose feature of a current frame to the appearance feature of {\it any} frame in the same video to reconstruct the same target, it enforces the appearance features to be similar across all frames.

%------------------------------------------------------------------------

\Paragraph{Gait Similarity Loss}
The cross reconstruction loss prevents the appearance feature $\fa$ to be over-represented, containing pose variation that changes between frames. 
However, appearance information may still be leaked into pose feature $\fg$. 
In an extreme case, $\fa$ is a constant vector while $\fg$ encodes all the information of a video frame.
To make $\fg$ ``cleaner", we leverage multiple videos of the same subject. Extra videos can introduce the change in appearance. Given two videos of the same subject with length $n_1$, $n_2$ in two different conditions $c_1$, $c_2$. Ideally, $c_1$, $c_2$ should contain difference in the person's appearance, i.e., cloth changes. While appearance changes, the gait information should be consistent between two videos. Since it's almost impossible to enforce similarity on $\fg$ between video frames as it requires precise frame-level alignment; we enforce the similarity between two videos' averaged pose features:
\begin{equation} \label{eq5}
\L_{\text{gait-sim}} =\norm{ \frac{1}{n_1}\sum_{t=1}^{n_1} \fg^{(t,c_1)} - \frac{1}{n_2}\sum_{t=1}^{n_2} \fg^{(t,c_2)} }^2_2.
\end{equation}

%-------------------------------------------------------------------------
\SubSection{Gait Feature Learning via Aggregation}
Even when we can disentangle appearance and pose information for each video frame, the current feature $\mathbf{f}_g$ only contains the walking pose of the person in a specific instance, which can share similarity with another specific instance of a very different person. 
Here, we are looking for discriminative characteristics in a person walking pattern. 
Therefore, modeling its temporal change is critical. This is where temporal modeling architectures like the recurrent neural network or long short-term memory (LSTM) work best.

Specifically, in this work, we utilize a multi-layer LSTM structure to explore spatial~(\eg, the shape of a person) and mainly, temporal~(\eg, how the trajectory of subjects' body parts changes over time) information on pose features. As shown in Fig.~\ref{fig-arc}, pose features extracted from one video sequence are feed into a $3$-layer LSTM.
The output of the LSTM is connected to a classifier $C$, in this case, a linear classifier is used, to classify the subject's identity. 

Let $\mathbf{h}^t$ be the output of the LSTM at time step $t$, which is accumulative after feeding $t$ pose features $\fg$ into it:
\begin{equation}
    \mathbf{h}^t = \text{LSTM} ( \fg^1,\fg^2, ..., \fg^t ).
\end{equation}

Now we define the loss function for LSTM.
A trivial option for identification is to add  the classification loss on top of the LSTM output of the final time step:
\begin{equation} \label{eqn:id_old}
\L_{\text{id-single}} = - \log( C_k( \mathbf{h}^n ) ),
\end{equation}
which is the negative log likelihood that the classifier $C$ correctly identifies the final output $\mathbf{h}^n$ as its identity label $k$.

\Paragraph{Identification with Averaged Feature}
By the nature of LSTM, the output $\mathbf{h}^t$ is greatly affected by its last input $\fg^t$. Hence the LSTM output, $\mathbf{h}^t$, can be varied across time steps. 
With a desire to obtain a gait feature that can be robust to the stopping instance of a walking cycle, we propose to use the averaged LSTM output as our gait feature for identification:
\begin{equation} \label{eqn:id}
\mathbf{f}_{\text{gait}}^t = \frac{1}{t} \sum_{s=1}^t \mathbf{h}^s.
\end{equation}
The identification loss can be rewritten as:
\begin{align} \label{eqn:id_averaged}
\L_{\text{id-avg}} & = - \log( C_k( \mathbf{f}_{\text{gait}}^n ) ) \nonumber \\
                   & = -\log\left( C_k\left( \frac{1}{n} \sum_{s=1}^n \mathbf{h}^s \right) \right).
\end{align}

\Paragraph{Incremental Identity Loss} 
LSTM is expected to learn that the longer the video sequence, the more walking information it processes then the more confident it identifies the subject. Instead of minimizing the loss on the final time step, we propose to use all the intermediate outputs of every time step weighted by $w_t$: %with more weight applied to later predictions: 
\begin{equation} \label{eqn:id}
\L_{ \text{id-inc-avg} } = \frac{1}{n}\sum_{t=1}^{n} - w_t \log \left( C_k \left( \frac{1}{t} \sum_{s=1}^t \mathbf{h}^s \right)\right).
\end{equation}

To this end, the overall training loss function is:
\begin{equation}
    \L = \L_{\text{id-inc-avg}} + \lambda_{r} \Lxrecon + \lambda_s  \L_{\text{gait-sim}}.
    \label{eqn:finalloss}
\end{equation}

The entire system, encoder-decoder, and LSTM are jointly trained. 
Updating $\Enc$ to optimize $\L_{\text{id-inc-avg}}$ also helps to further generate pose feature that has identity information and on which LSTM is able to explore temporal dynamics.
At the test time, the output $\mathbf{f}_{\text{gait}}^t$ of LSTM is the gait feature of the video and used as the identity feature representation for matching. The cosine similarity score is used as the metric.

\SubSection{Implementation Details}

\Paragraph{Segmentation and Detection}
Our network receives video frames with the person of interest segmented.
The foreground mask is obtained from the state-of-the-art instance segmentation, Mask R-CNN~\cite{he2017mask}.
Instead of using a zero-one mask by hard thresholding, we keep the soft mask returned by the network, where each pixel indicates the probability of being a person. 
This is partially due to the difficulty in choosing a threshold. 
Also, it prevents the loss in information due to the mask estimation error.
We use a bounding box with a fixed ratio of width : height $= 1: 2$ with the absolute height and center location given by the Mask R-CNN network.
Input is obtained by pixel-wise multiplication between the mask and RGB values which is then resized to $32\times64$.

\Paragraph{Network hyperparameter}
Our encoder-decoder network is a typical CNN. 
Encoder consisting of $4$ stride-2 convolution layers following by Batch Normalization and Leaky ReLU activation.
The decoder structure is an inverse of the encoder, built from transposed convolution, Batch Normalization and Leaky ReLU layers. 
The final layer has a Sigmoid activation to bring the value into $[0, 1]$ range as the input. 
The classification part is a stacked $3$-layer LSTM~\cite{gers1999learning}, which has $256$ hidden units in each of cells.

%To train the autoencoder and LSTM
Adam optimizer~\cite{kingma2014adam} is used with the learning rate of $0.0001$, and the momentum of $0.9$. %and batch size of $32$.
For each batch, we use video frames from $32$ different clips. 
Since video lengths are varied, a random crop of $20$-frame sequence is applied; all shorter videos are discarded. 
For Eqn.~\ref{eqn:id}, we set $w_t=t^2$ while other options such as $w_t=1$ also yield similar performance. The $\lambda_{r}$ and $\lambda_{s}$ (Eqn.~\ref{eqn:finalloss}) are set to $0.1$ and $0.005$ in all experiments.
%
%To minimize the loss in Eqn.~\ref{eqn:finalloss}, we use the Adam optimizer with $0.9$ and $0.99$ coefficients for computing running averages and squares of gradient~\cite{kingma2014adam}.

%-------------------------------------------------------------------------

\begin{figure}[t]
\includegraphics[width=\linewidth]{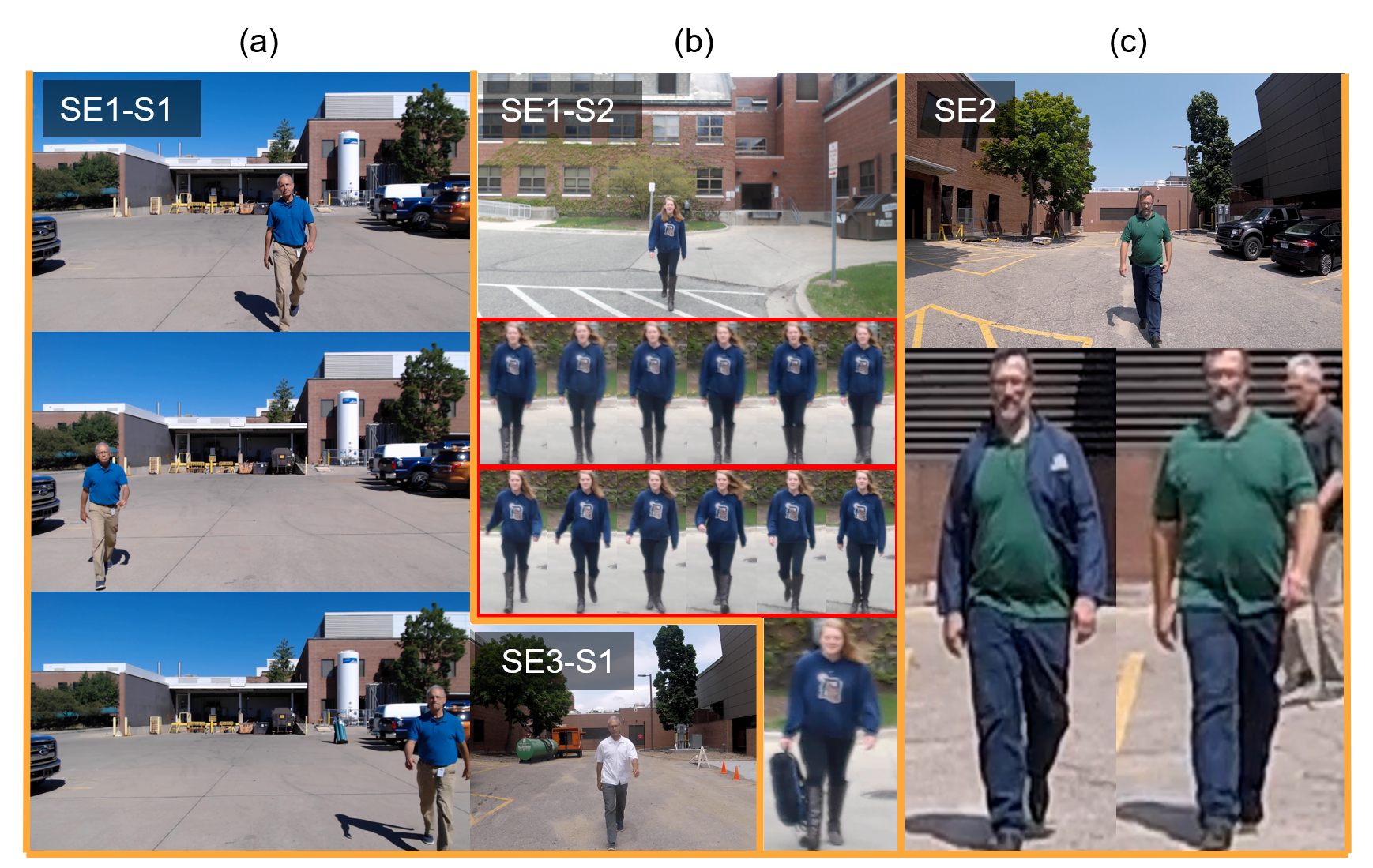}
\centering
\caption{Examples of FVG Dataset. (a) Samples of the near frontal middle, left and right walking view angles in session $1$ (SE$1$) of the first subject (S$1$). SE$3$-S$1$ is the same subject in session $3$. (b) Samples of slow and fast walking speed for another subject in session $1$. Frames in top red boxes are slow and in the bottom red box are fast walking. Carrying bag sample is shown below. (c) samples of changing clothes and with cluttered background from one subject in session $2$. }
\label{FIG-FVG}
\vspace{-5mm}
\end{figure}

\Section{Front-View Gait Database}

\Paragraph{Collection}
To facilitate the research of gait recognition from frontal-view angles, we collect the Front-View Gait (FVG) database in a course of two years 2017 and 2018.
During the capturing, we place the camera (Logitech C920 Pro Webcam or GoPro Hero 5) on a tripod at the height of $1.5$ meter. 
We ask each of $226$ subjects to walk toward the camera $12$ times starting from around $16$ meters, which results in $12$ videos per subject.
The videos are captured at $1,080\times1,920$ resolution with the average length of $10$ seconds.
The height of human in the video ranges from $101$ to $909$ pixels.
These $12$ walks have the combination of three angles toward the camera ($-45^\circ$, $0^\circ$,  $45^\circ$ off the optical axes of the camera), and four variations. 

FVG is collected in three sessions.
In session 1, in 2017, videos from $147$ subjects  are collected with four variations (normal walking, slow walking, fast walking, and carrying status). In session 2, in 2018, videos from additional $79$ subjects are collected. 
Variations are normal, slow or fast walking speed, clothes or shoes change, and twilight or clustered background. Finally in session 3, we collect repeated $12$ subjects in year $2018$ for extreme challenging test with the same setup as section 1. The purpose is to test how time gaps affect gait, along with changes in cloth/shoes or walking speed.
Fig.~\ref{FIG-FVG} shows exemplar images from FVG.

\Paragraph{Protocols}
Different from prior gait databases, subjects in FVG are walking toward the camera, which creates a great challenge on exploiting gait information as the difference in consecutive frames can be much smaller than side-view walking.
We focus our evaluation on variations that are challenging, \eg, different appearance, carrying a bag, or are not presented in other databases, \eg, cluttered background, along with view angles.

To benchmark research on FVG, we define $5$ evaluation protocols, among which there are two commonalities: $1)$ the first $136$ and rest $90$ subjects are used for training and testing respectively; $2)$ the video $2$, the normal frontal-view walking, is used as the gallery. 
The $5$ protocols differ in their specific probe data, which cover the variations of Walking Speed (WS), Carrying Bag (CB), Changing Clothes (CL), Cluttered Background (CBG), and all variations (All).
At the top part of Fig.~\ref{table-fvg}, we list the detailed probe set for all $5$ protocols.
\Eg, for the WS protocol, the probes are video $4-9$ in session $1$ and video $4-6$ in session $2$.  

\begin{figure}[t]
\includegraphics[width=8cm]{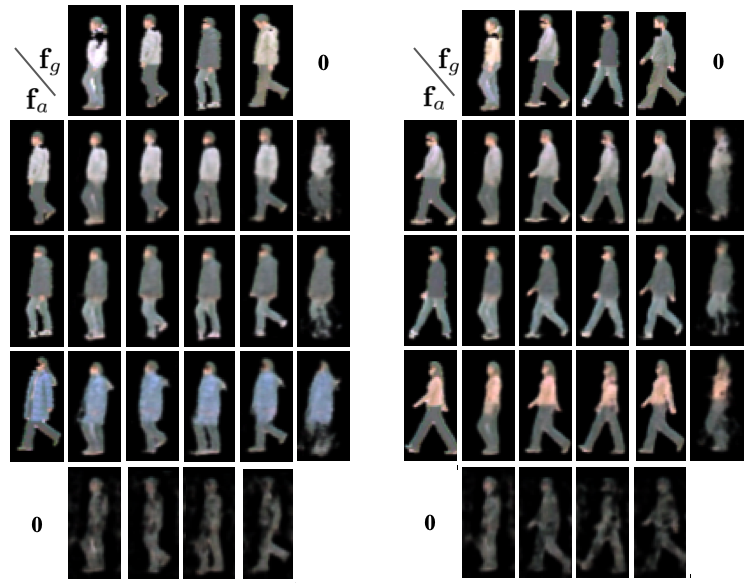}
\centering
\caption{Synthesized frames on CASIA-B by decoding the various combination of $\fa$ and $\fg$. Left and right parts are two examples. For each example, $\fa$ is extracted from images in the first column and $\fg$ is extracted from images in the first row. $\mathbf{0}$ vector has the same dimension as $\fg$ or $\fa$, accordingly.}
\label{fig:decodeimages}
\vspace{-5mm}
\end{figure}

\Section{Experiments}

\paragraph{Databases.}
We evaluate the proposed approach on three gait databases, CASIA-B~\cite{yu2006framework}, USF~\cite{sarkar2005humanid} and FVG.
As mentioned in Sec.~\ref{sec:relatedwork}, CASIA-B, and USF are the most widely used gait databases, making the comparison with prior work easier.
We compare our method with~\cite{wu2017comprehensive,chen2018multi,kusakunniran2010support,kusakunniran2014recognizing} on these two databases, by following the respective experimental protocols of the baselines. 
These are either the most recent and state-of-the-art work or classic gait recognition methods. 
The OU-ISIR database~\cite{makihara2012isir} is not evaluated, and related methods~\cite{makihara2017joint} are not compared since our work consumes RGB video input, but OU-ISIR only releases silhouettes.

%\vspace{-4mm}

%\begin{center}
%Example 1\space \space \space \space \space \space \space \space \space \space \space \space \space \space \space \space \space \space \space \space \space \space \space \space \space \space \space \space \space \space Example 2
%\end{center}

%\vspace{-4mm}

\begin{figure}[t]
\includegraphics[width=8cm]{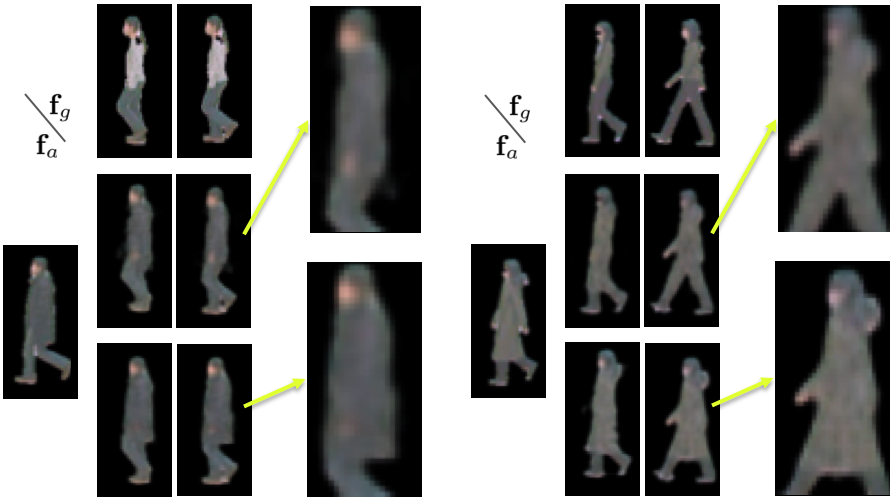}
\centering
\caption{Synthesized frames on CASIA-B by decoding $\fa$ and $\fg$ from different variations (NM vs.~CL). Left and right parts are two examples. For each example, $\fa$ is extracted from the most left column image (CL) and $\fg$ is extracted from the top row images (NM). Top row synthesized images are generated with model trained without $\L_{\text{gait-sim}}$ loss, bottom row is with the loss. To show the differences, details in generated images are magnified.}
\label{fig:decodewithGaitSimilarityLoss}
\vspace{-2mm}
\end{figure}

%\begin{table}[t]
%\caption{Ablation study on our disentanglement loss and classification loss. By removing or replacing with other loss functions, Rank-1 recognition rate on cross NM and CL condition degrades.}
%\label{table-ablation}
%\vspace{-2mm}
%\centering
%\resizebox{0.9\linewidth}{!}{
%\begin{tabular}{|l|l|c|}
%\hline
%Disentanglement Loss & Classification Loss & Rank $1$ \\ \hline
%- & $\Lid$ & $56.0$ \\
%$\Lxrecon $ & $\Lid$  & $60.2$   \\ \hline 
%$\Lxrecon + \L_{\text{gait-sim}}$ & $\Lid$ & $85.6$   \\ \hline
%$\Lxrecon + \L_{\text{gait-sim}}$ & $\L_{\text{id-avg}}$  & $62.6$   \\ %\hline
%%$\Lxrecon $ & $\L_{\text{id}}$  & 60.2   \\ \hline
% $\Lxrecon + \L_{\text{gait-sim}}$ & $\L_{\text{id-single}}$  & $26.0$   \\ %\hline
%$\Lxrecon + \L_{\text{gait-sim}}$ & $\L_{\text{id-ae}}$~\cite{srivastava2015unsupervised}        & $71.2$   \\ %\hline
%%$\Lxrecon$ & $\L_{\text{id-single}} & 24.6   \\ %\hline
%%-                    & $\L_{\text{id-single}} & 20.8      \\ \hline
%\hline
%\end{tabular}}
%\vspace{-4mm}
%\end{table}

\begin{table}[t]
\caption{Ablation study on our disentanglement loss and classification loss. By removing or replacing with other loss functions, Rank-1 recognition rate on cross NM and CL condition degrades.}
\label{table-ablation}
\vspace{-2mm}
\centering
\resizebox{0.9\linewidth}{!}{
\begin{tabular}{ccc}
\toprule
Disentanglement Loss & Classification Loss & Rank $1$ \\ \midrule
- & $\Lid$ & $56.0$ \\
$\Lxrecon $ & $\Lid$  & $60.2$   \\ \hline 
$\Lxrecon + \L_{\text{gait-sim}}$ & $\Lid$ & $85.6$   \\ \hline
$\Lxrecon + \L_{\text{gait-sim}}$ & $\L_{\text{id-avg}}$  & $62.6$   \\ %\hline
%$\Lxrecon $ & $\L_{\text{id}}$  & 60.2   \\ \hline
 $\Lxrecon + \L_{\text{gait-sim}}$ & $\L_{\text{id-single}}$  & $26.0$   \\ %\hline
$\Lxrecon + \L_{\text{gait-sim}}$ & $\L_{\text{id-ae}}$~\cite{srivastava2015unsupervised}  & $71.2$   \\ %\hline
%$\Lxrecon$ & $\L_{\text{id-single}} & 24.6   \\ %\hline
%-                    & $\L_{\text{id-single}} & 20.8      \\ \hline
\bottomrule
\end{tabular}}
\vspace{-4mm}
\end{table}

\SubSection{Ablation Study}

\begin{table*}[t]
\centering
\caption{Recognition accuracy cross views under NM on CASIA-B dataset. One single GaitNet module is trained for all the view angles. }
\label{table-view-angle}
\vspace{-2mm}
\resizebox{0.65\textwidth}{!}{
\begin{tabular}{lccccccccccc}
\toprule
Methods                                      & $0^\circ$  & $18^\circ$ & $36^\circ$        & $54^\circ$            & $72^\circ$   & $108^\circ$  & $126^\circ$ & $144^\circ$ & $162^\circ$ & $180^\circ$ & Average \\ \midrule
CPM~\cite{chen2018multi}                     & $13$ & $14$ & $17$        & $27$            & $62$  & $65$   & $22$  & $20$  & $15$  & $10$  & $24.1$   \\ %\hline
GEI-SVR~\cite{kusakunniran2010support}         & $16$ & $22$ & $35$        & $63$            & $95$   & $95$   & $65$  & $38$  & $20$  & $13$  & $42.0$      \\ %\hline
CMCC~\cite{kusakunniran2014spatiot}            & $18$ & $24$ & $41$        & $66$            & $96$   & $95$   & $68$  & $41$  & $21$  & $13$  & $43.9$   \\ %\hline
ViDP~\cite{hu2013view}                       & $8$ & $12$ & $45$        & $80$            & $\mathbf{100}$  & $\mathbf{100}$  & $81$  & $50$  & $15$  & $8$   & $45.4$   \\ %\hline
STIP+NN~\cite{kusakunniran2014recognizing}         & $-$  & $-$  & $-$         & $-$             & $84.0$   & $86.4$ & $-$   & $-$   & $-$   & $-$   & $-$    \\ %\hline
LB~\cite{wu2017comprehensive}    & $18$ & $36$ & $67.5$      & $93$            & $99.5$ & $99.5$  & $92$  & $66$  & $36$  & $18$  & $56.9$   \\ %\hline
L-CRF~\cite{chen2018multi}                   & $38$ & $\mathbf{75}$ & $68$        & $93$            & $98$   & $99$   & $\mathbf{93}$  & $67$  & $76$  & $39$  & $67.8$   \\ %\hline
GaitNet (ours)                                    & $\mathbf{68}$   &74    & $\mathbf{88}$ & $91$ & $99$     &$98$      &$84$     &$\mathbf{75}$     &$76$     &$\mathbf{65}$     & $\mathbf{81.8}$        \\ 
\bottomrule
\end{tabular}}
\vspace{-4mm}
\end{table*}
% ----------------------------------------------------------------------------------------

\Paragraph{Feature Visualization}
To aid on understanding our features, we randomly pair $\fa$, $\fg$ features from different images and visualize the resultant paired feature by feeding it into our learned decoder $\Dec$. 
As shown in Fig.~\ref{fig:decodeimages}, each result is generated by paring the appearance $\fa$ in the first column, and the pose $\fg$ in the first row. The synthesized images show that indeed $\fa$ contributes all the appearance information, \eg, cloth, color, texture, contour, as they are consistent across each row. Meanwhile, $\fg$ contributes all the pose information, \eg, position of hand and feet, which share similarity across columns.
We also visualize features $\fa$, $\fg$ {\it individually} by forcing the other feature to be a zero vector $\mathbf{0}$. Without $\fg$, the reconstructed image still shares appearance similarity with $\fa$ input but does not show a clear walking pose. Meanwhile, when removing $\fa$, the reconstructed image still mimics the pose of $\fg$'s input.

\begin{table}[t]
\centering
\caption{Comparison on CASIA-B with cross view and conditions. Three models are trained for NM-NM, NM-BG, NM-CL. Average accuracies are calculated excluding probe view angles.}
\label{table-pami17}
\vspace{-2mm}
\resizebox{0.48\textwidth}{!}{%
\begin{tabular}{l|ccccc|cccc}
\toprule
Gallery NM \#$1$-$4$ & \multicolumn{5}{c|}{$0^\circ$-$180^\circ$} & \multicolumn{4}{c}{$36^\circ$-$144^\circ$} \\ \cmidrule{1-1} \cmidrule{2-6} \cmidrule{7-10}
Probe NM \#5-6 & $0^\circ$ & $54^\circ$ & $90^\circ$ & $126^\circ$ & Mean & $54^\circ$ & $90^\circ$ & $126^\circ$ & Mean \\ \midrule
CCA~\cite{bashir2010cross} & $-$ & $-$ & $-$ & $-$ & $-$ & $66.0$ & $66.0$ & $67.0$ & $66.3$ \\ %\hline
ViDP~\cite{hu2013view} & $-$ & $64.2$ & $60.4$ & $65.0$ & $-$ & $87.0$ & $87.7$ & $89.3$ & $88.0$  \\ %\hline
% PE-LSTM &  &  &  &  &  &  &  &  &  \\ \hline
LB~\cite{wu2017comprehensive} & $82.6$ & $94.3$ & $87.4$ & $94.0$ & $89.6$ & $98.0$ & $98.0$ & $99.2$ & $98.4$ \\ %\hline
GaitNet (ours) & $\mathbf{91.2}$ & $\mathbf{95.6}$ & $\mathbf{92.6}$ & $\mathbf{96.0}$ & $\mathbf{93.9}$ & $\mathbf{99.1}$ & $\mathbf{99.0}$ & $\mathbf{99.2}$ & $\mathbf{99.1}$ \\ 
\bottomrule

%\toprule
Probe BG \#1-2 & $0^\circ$ & $54^\circ$ & $90^\circ$ & $126^\circ$ & Mean & $54^\circ$ & $90^\circ$ & $126^\circ$ & Mean \\ \midrule
LB-subGEI~\cite{wu2017comprehensive} & $64.2$ & $76.9$ & $63.1$ & $76.9$ & $70.3$ & $89.2$ & $84.3$ & $91.0$ & $88.2$ \\ %\hline
% Ours w/ BG subset &  &  &  &  &  &  &  &  &  \\ \hline
GaitNet (ours) & $\mathbf{83.0}$ & $\mathbf{86.6}$ & $\mathbf{74.8}$ & $\mathbf{85.8}$ & $\mathbf{82.6}$ & $\mathbf{90.0}$ & $\mathbf{85.6}$ & $\mathbf{92.7}$ & $\mathbf{89.4}$ \\

\bottomrule

%\toprule

Probe CL \#1-2 & $0^\circ$ & $54^\circ$ & $90^\circ$ & $126^\circ$ & Mean & $54^\circ$ & $90^\circ$ & $126^\circ$ & Mean \\ \midrule
LB-subGEI~\cite{wu2017comprehensive} & $37.7$ & $61.1$ & $54.6$ & $59.1$ & $53.1$ & $77.3$ & $74.5$ & $74.5$ & $75.4$ \\ %\hline
% Ours w/ CL subset & 42.1 & 70.7 & 70.6 & 69.4 &  &  &  &  &  \\ \hline
GaitNet (ours)   & $\mathbf{42.1}$ & $\mathbf{70.7}$ & $\mathbf{70.6}$ & $\mathbf{69.4}$ & $\mathbf{63.2}$ & $\mathbf{80.0}$  & $\mathbf{81.2}$ & $\mathbf{79.4}$ & $\mathbf{80.2}$\\ 
\bottomrule

\end{tabular}%
}
\vspace{-5mm}
\end{table}

%------------------------------------------------------------------------

\Paragraph{Disentanglement with Gait Similarity Loss}
With the cross reconstruction loss, the appearance feature $\fa$ can be enforced to represent static information that shares across the video. However, as discussed, the feature $\fg$ can be spoiled or even encode the whole video frame. Here we show the need for the gait similarity loss $\L_{\text{gait-sim}}$ on the feature disentanglement.
Fig.~\ref{fig:decodewithGaitSimilarityLoss} shows the cross visualization of two different models learned with and without $\L_{\text{gait-sim}}$. Without 
$\L_{\text{gait-sim}}$ the decoded image shares some appearance characteristic, \eg, cloth style, contour, with $\fg$. 
Meanwhile with $\L_{\text{gait-sim}}$, appearance better matches with $\fa$.

\Paragraph{Joints Location as Pose Feature}
In literature, there is a large amount of effort in human pose estimation~\cite{feng2016learning}. Aggregating joint locations over time could be a good candidate for gait features. Here we compare our framework with a baseline, named PE-LSTM, using pose estimation results as the input to the same LSTM as ours. Using state-of-the-art pose estimator~\cite{fang2017rmpe}, we extract $14$ joints' locations and feed to the LSTM. This network achieves the recognition accuracy of $65.4\%$ TDR at $1\%$ FAR on the ALL protocol of FVG dataset, where our method outperforms it with $81.2\%$.
This result demonstrates that our pose feature $\fg$ does explore more discriminate feature than the joints' locations alone.

\Paragraph{Loss Function's Impact on Performance} 
As the system consists of multiple loss functions, here we analyze the effect of each loss function on the final recognition performance. 
Tab.~\ref{table-ablation} reports the recognition accuracy of different variants of our framework on CASIA-B dataset under NM and CL. 
We first explore the effects of different disentanglement losses. Using $\Lid$ as the classification loss, we train different variants of our framework: a baseline without any disentanglement losses, a model with $\Lxrecon$, and our full model with both $\Lxrecon$ and $\L_{\text{gait-sim}}$. The baseline achieves the accuracy of $56.0\%$. Adding the $\Lxrecon$ slightly improves the performance to $60.2\%$. 
By combining with $\L_{\text{gait-sim}}$, our model significantly improves the performance to $85.6\%$.
Between $\Lxrecon$ and $\L_{\text{gait-sim}}$, the gait similarity loss plays a more critical role as $\Lxrecon$ is mainly designed to constrain the appearance feature $\fa$, which does not directly involve identification.

Using the combination, $\Lxrecon$ and $\L_{\text{gait-sim}}$, we benchmark different options for classification loss as presented in Sec.~\ref{sec:disentangle}, as well as the autoencoder loss by Srivastava et al.~\cite{srivastava2015unsupervised}.
The model using the conventional identity loss on the final LSTM output $\L_{\text{id-single}}$ achieves the rank-1 accuracy of $26.0\%$. 
Using the average output of LSTM as identity feature, $\L_{\text{id-average}}$, shows to improve the performance to $62.6\%$. The autoencoder loss~\cite{srivastava2015unsupervised}  achieves a good performance, $71.2\%$. However, it is still far from our proposed incremental identity loss $\Lid$'s performance.

\SubSection{Evaluation on Benchmark Datasets}
\vspace{3mm}
\SubSubSection{CASIA-B}
\label{sec:CASIA-B}

Since various experimental protocols have been defined on CASIA-B, for a fair comparison, we strictly follow the respective protocols in the baseline methods.
Following~\cite{wu2017comprehensive}, Protocol $1$ uses the first $74$ subjects for training and rest $50$ for testing, regarding variations of NM (normal), BG (carrying bag) and CL (wearing a coat) with crossing view angles of $0^\circ$, $54^\circ$, $90^\circ$, and $126^\circ$. Three models are trained for comparison in Tab.~\ref{table-pami17}.
For the detailed protocol, please refer to~\cite{wu2017comprehensive}. 
Here we mainly compare our performance to Wu et al.~\cite{wu2017comprehensive}, along with other methods~\cite{hu2013view}. 
Under multiple view angles and cross three variations, our method (GaitNet) achieves the best performance on all comparisons.

Recently, Chen et al.~\cite{chen2018multi} propose new protocols to unify the training and testing where only one single model is being trained for each protocol. 
Protocol $2$ focuses on walking direction variations, where all videos used are in NM. 
The training set includes videos of first $24$ subjects in all view angles. 
The rest $100$ subjects are for testing. 
The gallery is made of four videos at $90^\circ$ view for each subject. 
Videos from remaining view angles are the probe. 
The rank $1$ recognition accuracy are reported in Tab.~\ref{table-view-angle}.
Our GaitNet achieves the best average accuracy of $81.8\%$ across ten view angles, with significant improvement on extreme views.
\Eg, at view angles of $0^\circ$, and $180^\circ$, the improvement margins are $30\%$ and $26\%$ respectively. 
This shows that GaitNet learns a better view-invariant gait feature than other methods. 

\begin{table}[t!]
\centering
\caption{Comparison with \cite{chen2018multi} and \cite{wu2017comprehensive} under different walking conditions on CASIA-B by accuracies. One single GaitNet model is trained with all gallery and probe views and the two conditions. }
\vspace{-2mm}
\label{CB-TB3}
\resizebox{0.47\textwidth}{!}
{%
\begin{tabular}{@{}cc@{}ccc@{}ccc@{}ccc@{}ccc@{}}
\toprule
\multirow{2}{*}{Probe} & \multirow{2}{*}{Gallery} && \multicolumn{2}{c}{GaitNet (ours)} && \multicolumn{2}{c}{L-CRF \cite{chen2018multi}} && \multicolumn{2}{c}{LB \cite{wu2017comprehensive}} && \multicolumn{2}{c}{RLTDA \cite{hu2013enhanced}} \\ \cmidrule{4-5} \cmidrule{7-8} \cmidrule{10-11} \cmidrule{13-14}
 &  && BG & CL && BG & CL && BG & CL && BG & CL \\ \midrule
$54$ & $36$ && $91.6$ & $\mathbf{87.0}$ && $\mathbf{93.8}$ & $59.8$ && $92.7$ & $49.7$ && $80.8$ & $69.4$ \\ 
$54$ & $72$ && ${90.0}$ & $\mathbf{90.0}$ && $\mathbf{91.2}$ & $72.5$ && $90.4$ & $62.0$ && $71.5$ & $57.8$ \\
$90$ & $72$ && $\mathbf{95.6}$ & $\mathbf{94.2}$ && $94.4$  & $88.5$ && $93.3$ &  $78.3$ && $75.3$ &  $63.2$ \\
$90$ & $108$ && $87.4$ & $\mathbf{86.5}$ && $\mathbf{89.2}$ & $85.7$ && $88.9$ & $75.6$ && $76.5$ &$72.1$ \\  
$126$ & $108$ && $90.1$ & $\mathbf{89.8}$ && $\mathbf{92.5}$ & $68.8$ && $93.3$ & $58.1$ && $66.5$ & $64.6$ \\
$126$ & $144$ && $\mathbf{93.8}$ & $\mathbf{91.2}$ && $88.1$ & $62.5$ && $86.0$ & $51.4$ && $72.3$ & $64.2$ \\
\midrule
\multicolumn{2}{c}{Mean}  &&  $91.4$ & $\mathbf{89.8}$ && $\mathbf{91.5}$ & $73.0$ && $90.8$ & $62.5$ && $73.8$ & $65.2$
\\ \bottomrule
\end{tabular}%
}
\end{table}

% ----------------------------------------------------------------------------------------

%
Protocol $3$ focuses on appearance variations. 
Training sets have videos under BG and CL. There are $34$ subjects in total with  $54^\circ$ to $144^\circ$ view angles. 
Different test sets are made with the different combination of view angles of the gallery and probe as well as the appearance condition (BG or CL). 
The results are presented in Tab.~\ref{CB-TB3}. 
We have comparable performance with the state-of-the-art method L-CRF~\cite{chen2018multi} on BG subset while significantly improving the performance on CL subset.
Note that due to the challenge of CL protocol, there is a significant performance gap between BG and CL for all methods except ours, which is yet another evidence that our gait feature has strong invariance to all major gait variations.

Across all evaluation protocols, GaitNet consistently outperforms state of the art. This shows the superior of GaitNet on learning a robust representation under different variations. It is contributed to our ability to disentangle pose/gait information from other static variations.

\begin{table}[t]
\centering
\caption{Definition of FVG protocols and performance comparison. 
Under each of the $5$ protocols, the first/second columns indicate the indexes of videos used in gallery/probe.}
\label{table-fvg}
% \vspace{1mm}
\vspace{-2mm}
\resizebox{0.47\textwidth}{!}{%
\begin{tabular}{l|cc|cc|cc|cc|cc}
\toprule
\multicolumn{11}{c}{Index of Gallery \& Probe videos} \\ \midrule
Session 1      &$2$ & $4$-$9$    & $2$ & $10$-$12$  & $-$ & $-$             & $-$ & $-$                       & $2$ & $1$,$3$-$12$ \\ 
Session 2      &$2$ & $4$-$6$          & $-$ & $-$             & $2$ & $7$-$9$     & $2$ & $10$-$12$                & $2$ & $1$,$3$-$12$ \\ 
Session 3      &$-$ & $-$                  & $-$ & $-$             & $-$ & $-$             & $-$ & $-$               & $-$ & $1$-$12$ \\ \midrule
Variation    & \multicolumn{2}{c|}{WS} & \multicolumn{2}{c|}{CB} & \multicolumn{2}{c|}{CL} & \multicolumn{2}{c|}{CBG} & \multicolumn{2}{c}{All} \\ \midrule
%\#Train/Test & \multicolumn{10}{c}{True Detection Rate (TDR) @1\% or 5\% False Alarm Rate (FAR)}             \\ \midrule
TDR@FAR      & $1\%$       & $5\%$      & $1\% $       & $5\%$        &$1\%$       & $5\%$      & $1\%$          & $5\%$          & $1\%$       & $5\%$       \\ \midrule
PE-LSTM      &      $79.3$        &       $87.3$      &       $59.1$        &       $78.6$        &        $55.4$      &        $67.5$     &        $61.6$         &        $72.2$         &      $65.4$        &      $74.1$        \\
GEI~\cite{han2006individual}  &    $9.4$     &      $19.5$        &       $6.1$      &       $12.5$        &         $5.7$      &       $13.2$       &      $6.3$       &          $16.7$       &     $5.8$       &       $16.1$   \\
% GEnI\cite{bashir2009gait}         &              &             &               &               &              &             &                 &                 &              &              \\ \hline
GEINet~\cite{shiraga2016geinet} & $15.5$     &      $35.2$        &       $11.8$      &     $24.7$          &         $6.5$      &     $16.7$         &      $17.3$       &        $35.2$        &       $13.0$       &       $29.2$       \\
DCNN~\cite{alotaibi2017improved} &  $11.0$      &       $23.6$       &     $5.7$       &       $12.7$        &       $7.0$        &       $15.9$    &      $8.1$           &        $20.9$         &       $7.9$       &     $19.0$         \\
LB~\cite{wu2017comprehensive}   &        $53.4$      &      $73.1$       &      $23.1$         &        $50.3$       &      $23.2$        &     $38.5$        &       $56.1$          &        $74.3$         &      $40.7$        &       $61.6$       \\ 
% MT Network\cite{wu2017comprehensive}   &              &             &               &               &              &             &                 &                 &              &              \\ \hline
% 3D-CNN Network\cite{wu2017comprehensive}   &              &             &               &               &              &             &                 &                 &              &              \\ \hline
GaitNet (ours) &  $\mathbf{91.8}$ &  $\mathbf{96.6}$ & $\mathbf{74.2}$ & $\mathbf{85.1}$ & $\mathbf{56.8}$ & $\mathbf{72.0}$ &  $\mathbf{92.3}$  &  $\mathbf{97.0}$ & $\mathbf{81.2}$ & $\mathbf{87.8}$       \\
\bottomrule
\end{tabular}%
}
\vspace{-4mm}
\end{table}

\SubSubSection{USF}
The original protocol of USF~\cite{sarkar2005humanid} does not define a training set, which is not applicable to our method, as well as ~\cite{wu2017comprehensive}, that require data to train the models. 
Hence following the experiment setting in~\cite{wu2017comprehensive}, we randomly partition the dataset into the non-overlapping training and test sets, each with half of the subjects.
We test on Probe A, defined in~\cite{wu2017comprehensive}, where the probe is different from the gallery by the viewpoint. We achieve the identification accuracy of $99.5\pm 0.2\%$, which is better than the reported $96.7\pm0.5\%$ of LB network~\cite{wu2017comprehensive}, and $94.7\pm2.2\%$ of multi-task GAN~\cite{he2019multi}.

\SubSubSection{FVG}
\label{sec:fvg}

Given that FVG is a newly collected database and no reported performance from prior work, we make the efforts to implement $4$ classic or state-of-the-art methods on gait recognition~\cite{han2006individual,shiraga2016geinet,alotaibi2017improved,wu2017comprehensive}.
For each of $4$ methods and our GaitNet, one model is trained with the $136$-subject training set and tested on all $5$ protocols.

As shown in Tab.~\ref{table-fvg}, our method shows state-of-the-art performance compared with other methods, including the recent CNN-based methods. Among $5$ protocols, CL is the most challenging variation as in CASIA-B. Comparing with all different methods GEI based methods suffer from frontal view due to the lack of walking information. 

\SubSection{Runtime Speed}

System efficiency is an essential metric for many vision systems including gait recognition. We calculate the efficiency while each of the $5$ methods processing one video of USF dataset on the same desktop with GeForce GTX 1080 Ti GPU. As shown in Tab.~\ref{runtime-table}, our method is significantly faster than the pose estimation method because of 1) efficiency of Mask R-CNN; 2) an accurate, yet slow, version of AlphaPose~\cite{fang2017rmpe} is required for gait recognition.

\begin{table}[t!]
\centering
\caption{Runtime (ms per frame) comparison on FVG dataset.}
\label{runtime-table}
\vspace{-2mm}
\resizebox{0.75\linewidth}{!}{%
\begin{tabular}{lccc}
\toprule
Methods              & Pre-processing & Inference & Total \\  \midrule
% Identical graphlets  & CPU    &      $124$     \\ \hline
% Participant tracking & CPU    &      $118$     \\ \hline
PE-LSTM              & $22.4$    &      $0.1$ &    $22.5$      \\
GEINet~\cite{shiraga2016geinet} & $0.5$    &      $1.5$  & $2.0$     \\ 
DCNN~\cite{alotaibi2017improved} & $0.5$    &      $1.7$  & $2.2$     \\
LB~\cite{wu2017comprehensive} & $0.5$    &      $1.3$ & $1.8$      \\ 
GaitNet (ours)                   & $0.5$    &       $1.0$  & $1.5$     \\
\bottomrule
\end{tabular}}
\vspace{-4mm}
\end{table}

\Section{Conclusions}
This paper presents an autoencoder-based method termed GaitNet that can disentangle appearance and gait feature representation from raw RGB frames, and utilize a multi-layer LSTM structure to further explore temporal information to generate a gait representation for each video sequence. 
We compare our method extensively with the state of the arts on CASIA-B, USF, and our collected FVG datasets. 
The superior results show the generalization and promising of the proposed feature disentanglement approach.
We hope that in the future, this disentanglement approach is a viable option for other vision problems where motion dynamics needs to be extracted while being invariant to confounding factors, \eg, expression recognition with invariance to facial appearance, activity recognition with invariance to clothing.
%It shows promising results on the frontal walking scenario, which is an essential and complex part of gait recognition and it is real life applications. Not only on frontal view, GaitNet generally works well on all varieties gait recognition including view angle, walking speed, carrying condition, etc. due to utilizing maximums information from RGB.

\vspace{-0.06in}
\subsubsection*{Acknowledgement}
\vspace{-0.05in}
\noindent This work was supported with funds from the Ford-MSU Alliance program.

{\small
\bibliographystyle{ieee_fullname}
\bibliography{egbib}
}

\end{document}